\documentclass[11pt]{article}

\usepackage[final]{acl}

\usepackage{times}
\usepackage{latexsym}

\usepackage[T1]{fontenc}

\usepackage[utf8]{inputenc}

\usepackage{microtype}

\usepackage{inconsolata}

\usepackage{graphicx}

\usepackage{amsmath}
\usepackage{amsthm}
\usepackage{amssymb}

\usepackage{multirow}
\usepackage{booktabs}
\usepackage{array}
\usepackage{soul}
\usepackage[table]{xcolor}
\usepackage{makecell} 
\usepackage{subfigure}
\usepackage{enumitem}

\usepackage{titletoc}    
\usepackage{hyperref}    
\usepackage{cleveref}   
\usepackage{fontawesome5} 
\usepackage{seqsplit}

\usepackage{helvet}

\definecolor{Gray}{RGB}{242,242,242}

\title{Personalizing LLMs with Binary Feedback: \\A Preference-Corrected Optimization Framework}

\author{
    Xilai Ma$^1$, Liye Zhao$^2$, Weijun Yao$^2$, Haibing Di$^2$, Wenya Wang$^3$,  \textbf{Jing Li}$^1$\textsuperscript{\texorpdfstring{\faIcon[regular]{envelope}}{}} 
    \\$^{1}$Harbin Institute of Technology, Shenzhen, China 
    \\$^{2}$Huawei Technologies Co., Ltd.  \quad $^{3}$Nanyang Technological University \\
    \texttt{maxilai.hour1@gmail.com} \quad \texttt{jingli.phd@hotmail.com}  
}

\begin{document}
\maketitle
\begin{abstract}

Large Language Model (LLM) personalization aims to align model behaviors with individual user preferences.
Existing methods often focus on isolated user histories, neglecting the essential role of inter-user differences.
We propose C-BPO, a framework that personalizes LLMs via preference-calibrated binary signals.
By treating target user data as positive feedback and other users' data as an auxiliary set of implicit negative signals, C-BPO captures distinct inter-user differences.
To mitigate the preference overlap issue, where shared task knowledge is erroneously penalized, we derive an objective grounded in Positive-Unlabeled (PU) learning theory.
This approach purifies negative signals by subtracting ``positive bias'', ensuring alignment with unique idiosyncrasies without compromising general helpfulness.
Empirical experiments across various personalization tasks and backbone LLMs show C-BPO consistently outperforms baselines, demonstrating the efficacy of preference-calibrated binary signals in modeling inter-user differences.
\let\thefootnote\relax\footnotetext{\faIcon[regular]{envelope}~Corresponding author.}
\end{abstract}

\section{Introduction}
\label{sec:intro}

The advent of Large Language Models (LLMs) has driven the realization of advanced applications such as question answering, planning, and interactive agents~\cite{ouyang2022training, touvron2023llama, qwen2}. These models not only excel at executing specific tasks guided by textual instructions, but also demonstrate exceptional potential in complex reasoning~\cite{shao2024deepseekmath, guo2026backdoors} and tool-assisted decision-making~\cite{zhang2025landscape, guo2026e3}.
However, most efforts to enhance LLM helpfulness adhere to a ``one-size-fits-all'' paradigm, optimizing for the preferences of an average user.
Consequently, LLM personalization has emerged as a pivotal research direction~\cite{salemi2024lamp, chen2024large, tseng2024two, guan-etal-2025-survey, li2025multi}, seeking to align model responses with individual user preferences.

Early prompt-based techniques~\cite{mysore2024pearl, richardson2023integrating} facilitate personalization by appending user-specific context, such as retrieved historical snippets or summarized user profiles, to the input query.
Furthermore, parameter-efficient fine-tuning (PEFT) methods have been introduced to learn lightweight, user-specific modules directly from historical data~\cite{tan2024democratizing, liu2025llms}, with further research exploring collaborative generation through multi-LoRA architectures~\cite{tan2024personalized, zhang2025proper}.
Despite their success, these methods primarily focus on the target user's isolated history.
Drawing from behavioral science~\cite{snyder2012uniqueness, irmak2010you}, where individuality is defined by inter-user variability, recent studies emphasize that effective personalization must capture how a user deviates from the general population. For instance, frameworks such as DPL~\cite{qiu-etal-2025-measuring} and DEP~\cite{qiu2025latent} have attempted to model these distinctions through either LLM-based textual comparisons or latent space embeddings of historical data.
However, these methods face scalability and granularity constraints: textual analyzers are limited by the high computational cost of LLM reasoning, while latent embeddings often fail to capture fine-grained, phrase-level stylistic nuances.


While preference optimization frameworks like DPO~\cite{rafailov2023direct} offer a solution, they require contrastive completions ($y_{win}, y_{lose}$) for identical inputs $x$, which are unavailable in individualized user historical data.
Furthermore, such frameworks lack a mechanism to exploit inherent inter-user information for refined preference learning.
This leads to a fundamental question: \textit{Can we model individual preferences directly from raw, unpaired data by leveraging inter-user distinctions?}

Inspired by recent advances~\cite{ethayarajh2024model, jung2025binary} in preference optimization with binary feedback, it has been demonstrated that LLMs can be aligned using binary signals, such as ``thumbs-up'' or ``thumbs-down'' labels assigned to individual data points $(x, y)$. This approach circumvents the requirement for shared inputs $x$ and contrastive completions $(y_{win}, y_{lose})$ in traditional preference datasets.
In this work, we argue that the target user's historical data can be treated as positive feedback, while data from other users serves as implicit negative feedback.
This formulation allows the model to capture inter-user distinctions directly within a binary-feedback preference optimization framework. 
However, existing binary-feedback preference optimization (BPO) methods are typically designed for ``clean-labeled preference'' where negative samples are objectively inferior according to average human preferences.
In personalized contexts, naively treating other users' data as purely negative signals incurs a \textbf{preference overlap} issue (as detailed in \S~\ref{sec:motivation}). 
Specifically, this leads to the excessive penalization of \textbf{generic task-specific knowledge} and \textbf{ community-wide preferences} that are inherently shared between the target user and the broader population.

To adapt BPO to personalized scenarios, we derive a preference-calibrated objective grounded in Positive-Unlabeled (PU) learning theory~\cite{bekker2020learning}.
By leveraging the core concept of unbiased risk decomposition, we reformulate the optimization objective to account for the fact that data from other users acts as an unlabeled mixture of both common and user-specific features (as detailed in \S~\ref{sec:pu}).
Specifically, we derive an estimator that recovers the true negative risk by subtracting the ``positive bias'' from the total auxiliary risk, where this bias represents the expected risk of target preferences found within the broader user data.
This objective explicitly purifies the negative signals, ensuring that the model focuses on unique individual idiosyncrasies rather than suppressing general helpfulness.
Furthermore, to handle the data imbalance between these two categories of data, we introduce an independent exponential moving average (EMA) reference point estimation to maintain a stable preference boundary during training (as detailed in \S~\ref{sec:imbalance}).


In summary, we propose C-BPO, a framework designed for personalizing LLMs through preference-calibrated binary-feedback signals.
To evaluate its effectiveness, we conduct extensive experiments across five personalized generation tasks using various LLMs. 
Our contributions are summarized as follows:
\begin{itemize}[leftmargin=*, noitemsep, nolistsep]

    \item To the best of our knowledge, we are the first to frame LLM personalization as a binary-feedback preference optimization (BPO) problem, utilizing inter-user variability as a natural contrastive signal for individual-level alignment.

    \item We uncover the \textbf{preference overlap} issue in standard BPO for personalization (\S~\ref{sec:motivation}) and formulate a preference-calibrated objective grounded in Positive-Unlabeled (PU) learning (\S~\ref{sec:pu}). 
    
    \item Extensive experiments demonstrate that C-BPO consistently outperforms strong baselines.
    We also provide in-depth analyses of how sample volume (\S~\ref{sec:exp_aux}) and preference overlap (\S~\ref{sec:exp_unique}) influence the optimization dynamics.
\end{itemize}


\section{Preliminaries}
\label{sec:preliminaries}

To align large language models with general human preferences (e.g., helpfulness and harmlessness), representative approaches such as Reinforcement Learning from Human Feedback (RLHF, ~\citealp[]{ouyang2022training}) and Direct Preference Optimization (DPO, ~\citealp[]{rafailov2023direct}) typically rely on paired preference data to explicitly or implicitly construct reward signals.

\paragraph{DPO.}
\citet{rafailov2023direct} shows that the policy $\pi_\theta$ can be directly optimized from the paired preference dataset $\mathcal{D}$, and the implicit reward function can be defined as a function of the policy:
\begin{equation}\label{eq:reward}
r_\theta(x,y) = \beta \log \frac{\pi_\theta(y \mid x)}{\pi_{\text{ref}}(y \mid x)}
\end{equation}
Combining the BT model with the implicit reward, the loss function of DPO is
\begin{align*}
    -\mathbb{E}_{(x,y_w,y_l) \sim \mathcal{D}} \left [ \log \sigma \left( r_\theta(x,y_w) - r_\theta(x,y_l)\right) \right].
\end{align*}
Here, $y_w$ is a preferred completion and $y_l$ is a non-preferred completion.

\paragraph{Preference Optimization with Binary Feedback.}
To overcome the limitation of RLHF that human feedback is provided as pairwise preferences over multiple outputs for the same input $x$ (e.g., $y_w > y_l$ for the same input $x$), recent works~\citep{ethayarajh2024model, jung2025binary} have shown that LLMs can be aligned using binary feedback, where ``thumbs-up'' or ``thumbs-down'' signals are assigned to individual $(x,y)$ data point--completion pairs without requiring shared inputs.
KTO~\citep{ethayarajh2024model} and BCO~\cite{jung2025binary} are two representative methods in this line of work.
Both of them optimize a closely related objective under the implicit reward formulation induced by the policy--reference relationship in Eq.~\eqref{eq:reward}.
We take the BCO objective as an illustrative example\footnote{More detailed comparison are given in \S~\ref{appendix:bco_comparison}}:
\begin{align}
&\mathbb{E}_{(x, y_w) \sim \mathcal{D}^+}
\left[- \log \sigma\!\left(r_\theta(x, y_w) - \delta\right)\right] \nonumber \\
&\quad + \mathbb{E}_{(x, y_l) \sim \mathcal{D}^-}
\left[- \log \sigma\!\left(- (r_\theta(x, y_l) - \delta)\right)\right],
\label{eq:bco_objective}
\end{align}
where $\mathcal{D}^+$ and $\mathcal{D}^-$ denote ``thumbs-up'' and ``thumbs-down'' datasets, respectively, and $\delta$ is a reference point that anchors the implicit reward.

The main difference between KTO and BCO lies in how they define the reference point $\delta$ in Eq.~\eqref{eq:bco_objective}.
KTO sets $\delta$ as the average implicit reward of \emph{other completions} in the same batch, with zero-clipping to ensure non-negativity (as shown in Eq.~\eqref{eq:kto_z_ref}).

In contrast, BCO improves upon KTO by taking the average implicit reward of contrastive pairs from the training batch as the reference point:
\begin{equation}
\begin{split}
\delta_{\text{BCO}} = \frac{1}{2} \Big( &\mathbb{E}_{(x,y)\sim \mathcal{D}^+}[r_\theta(x,y)] \\
&+ \mathbb{E}_{(x,y)\sim \mathcal{D}^-}[r_\theta(x,y)] \Big).
\end{split}
\end{equation}
This modification reduces bias introduced by batch-based reference points and preserves informative gradients for all samples, enabling more stable training purely from binary feedback.

\newtheorem{lemma}{Lemma}
\newtheorem{definition}{Definition}
\newtheorem{proposition}{Proposition}

\newcommand{\E}{\mathbb{E}}
\newcommand{\D}{\mathcal{D}}
\newcommand{\Lcal}{\mathcal{L}}
\newcommand{\R}{\mathbb{R}}
\newcommand{\loss}{\ell}

\section{Personalization with Binary-Feedback Preference Optimization}
\label{sec:personalization_pc_bfo}

\begin{figure}[t!]
  \centering
   \includegraphics[width=1.0\linewidth]{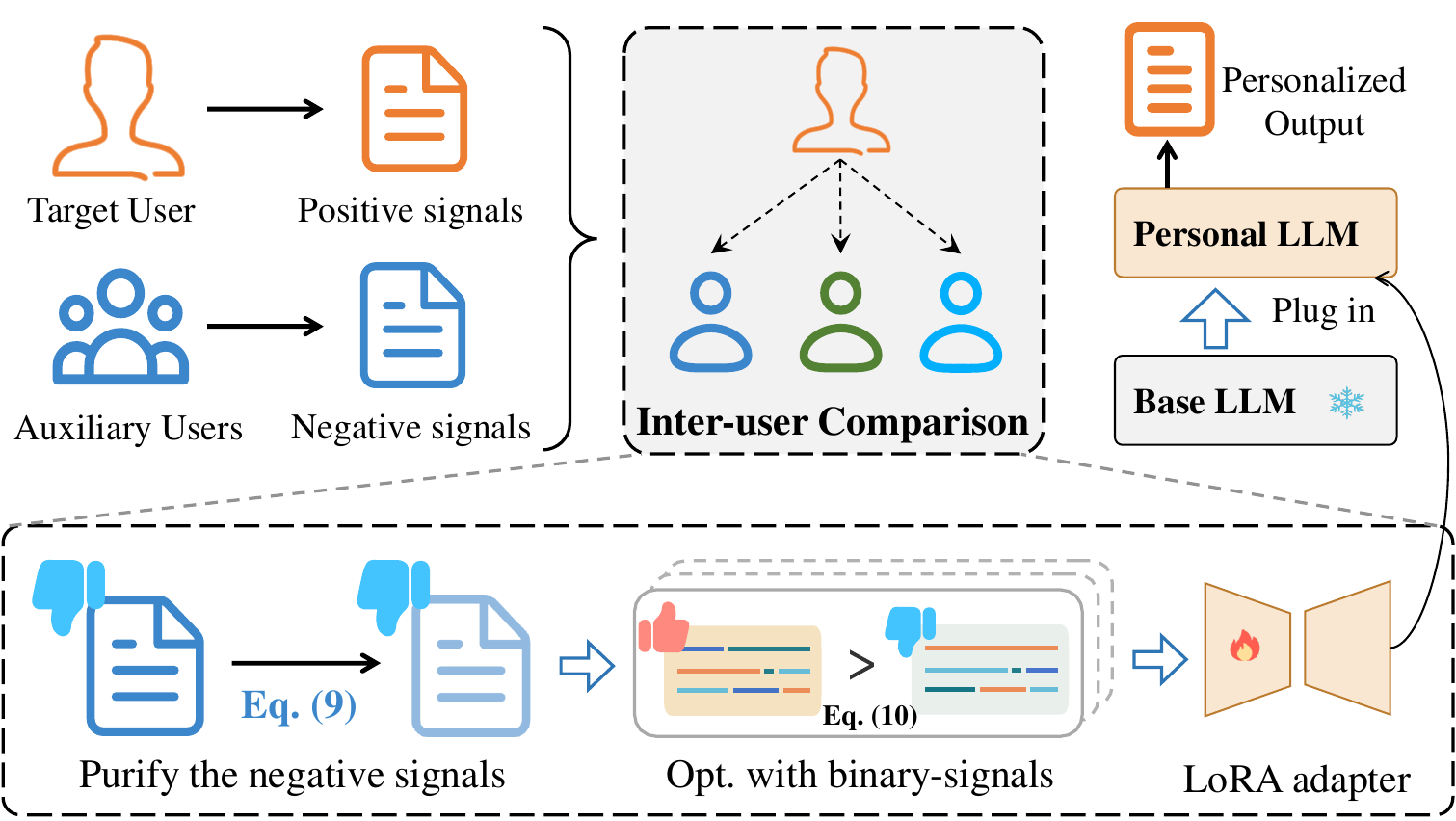}
   \caption{Overview of the \textbf{C-BPO} framework. We leverage the target user's data as positive signals and auxiliary users' data as implicit negative signals to align the LLM with the target user's distinct preferences.}
   \label{fig:overview}
\end{figure}

In this section, we introduce C-BPO, a framework designed for personalizing LLMs through preference-calibrated binary-feedback signals (Figure~\ref{fig:overview}).
We first formulate the personalization task within a binary-feedback preference optimization framework and analyze the potential biases arising from ``negative data'' in \S\ref{sec:motivation}.
Subsequently, in \S\ref{sec:pu}, we derive a preference-corrected optimization objective based on Positive-Unlabeled (PU) learning theory to achieve unbiased preference alignment.
Finally, in \S\ref{sec:imbalance}, we discuss how to extend our framework to scenarios characterized by data imbalance between user-specific and ``negative'' sets.

\subsection{Problem Formulation and Motivation}
\label{sec:motivation}

The goal of LLM personalization is to adapt a pre-trained model $\pi_{\text{base}}$ to align with the specific preferences and behaviors of a target user, which are characterized by their historical interaction data $\mathcal{H}_{\text{user}} = \{(x_i, y_i)\}_{i=1}^N$.
A standard practice for efficient adaptation is to employ Parameter-Efficient Fine-Tuning (PEFT) techniques, such as LoRA~\cite{hulora}, to optimize a user-specific module $\Delta_{\text{user}}$.
The personalized model is thus defined as $\pi_{\text{user}} = \pi_{\text{base}} + \Delta_{\text{user}}$. Conventional approaches typically optimize $\Delta_{\text{user}}$ via supervised fine-tuning (SFT) on $\mathcal{H}_{\text{user}}$, minimizing the cross-entropy loss between the model's output and the historical ground-truth labels.
While standard practices rely solely on $\mathcal{H}_{\text{target}}$ for supervised training, we investigate the following question: \textit{can we directly capture inter-user distinctions by modeling preferences within the raw historical data? }


Our key motivation stems from \textbf{Binary-Feedback Preference Optimization} (\S\ref{sec:preliminaries}), which facilitates preference alignment without requiring explicit pairwise signals.
Under this framework, we can treat the \textbf{target user's data $\mathcal{H}_{\text{tar}}$} as positive feedback (``thumbs-up'') and the data from \textbf{auxiliary users $\mathcal{H}_{\text{aux}}$}\footnote{Throughout this paper, we refer to the user undergoing optimization as the target user and a selected pool of other users as auxiliary users.} as implicit negative feedback (``thumbs-down'').
This allows us to directly capture inter-user difference signals from raw data by employing the Eq.~\eqref{eq:bco_objective}.

While existing binary-feedback preference optimization methods typically assume ``clean scenarios'', where $(x, y_w)\in\mathcal{D}^+$ and $(x, y_l)\in\mathcal{D}^-$ are explicitly labeled as preferred and non-preferred.
Naively treating $\mathcal{H}_{\text{aux}}$ as the negative set $\mathcal{D}^-$ introduces a critical challenge: preference overlap.
Specifically, different user data inevitably share common preferences~\cite{zhang2025proper}, encompassing \textbf{generic task-specific knowledge} defined by the core requirements of a particular task, and \textbf{community-wide preferences} characterized by collective stylistic or aesthetic trends shared across the population.
Directly penalizing $\mathcal{H}_{\text{aux}}$ via Eq. \eqref{eq:bco_objective} forces the model to erroneously suppress shared features, impeding effective individual-level preference alignment.
To mitigate this issue, we aim to ``peel off'' the common preference signals from the noisy auxiliary data, ensuring a more accurate and robust personalization.


\subsection{Unbiased Personalization via PU Risk Reformulation}
\label{sec:pu}

To address the negative bias discussed in \S\ref{sec:motivation}, we propose to reformulate the binary-feedback preference optimization under Positive-Unlabeled (PU) learning theory~\cite{kiryo2017positive, bekker2020learning, wang2023pue}.

\paragraph{Unbiased Risk Estimation.}
In standard binary classification, let $L(g(x), y)$ be a loss function for a classifier $g$. The expected risk $R(g)$ is defined as:
\begin{equation}
    R(g) = \pi_p R^+_p(g) + \pi_n R^-_n(g),
    \label{eq:standard_risk}
\end{equation}
where $\pi_p/\pi_n$ is the positive/negative class prior, and $R^+_p(g)=\mathbb{E}_{x \sim p_p}[L(g(x), +1)]$ and $R^-_n(g) = \mathbb{E}_{x \sim p_n}[L(g(x), -1)]$ denote the expected positive and negative risks, respectively.

In standard PU learning, the negative distribution $p_n$ is likewise unavailable; instead, we utilize a set of unlabeled samples drawn from the marginal density $p_u(x) = \pi_p p_p(x) + \pi_n p_n(x)$.
This relationship allows for the following risk decomposition:

\begin{lemma}[Risk Decomposition] \label{lem:risk_decomp_main}
Let $p_u(x) = \pi_p p_p(x) + \pi_n p_n(x)$ be the marginal distribution of unlabeled data. The negative risk can be unbiasedly estimated using only positive and unlabeled data distributions:
\begin{equation}
    \pi_n R^-_n(g) = R^-_u(g) - \pi_p R^-_p(g).
    \label{eq:pu_equality}
\end{equation}
\end{lemma}

The proof and further details on PU Learning are provided in Appendix~\ref{sec:proofs} and \ref{sec:pu_learning}, respectively.
Eq.~\eqref{eq:pu_equality} suggests that the negative bias (the ``misclassified'' positive component) inherent in $p_u$ can be explicitly subtracted to recover the true underlying negative signal and ensure a stable estimation.

\paragraph{C-BPO Objective.}
Under the binary-feedback preference optimization framework, we define $g$ as a discriminator capable of distinguishing between preferred and non-preferred samples, which in practice is parameterized by the user-specific module $\Delta_{\text{user}}$ to capture preference boundaries.
Based on the binary signals introduced in \S~\ref{sec:preliminaries}, we formulate the following preference loss functions:
\begin{align}
    l(g(x, y), +1) &= -\log \sigma(r_\theta(x, y) - \delta), \\
    l(g(x, y), -1) &= -\log \sigma(-(r_\theta(x, y) - \delta)),
\end{align}
where $r_\theta(x, y)$ is derived from Eq.~\eqref{eq:reward}.
Here, $l(g, +1)$ quantifies the degree to which a sample is preferred, whereas $l(g, -1)$ measures the extent to which it is non-preferred.

Following Eq.~\eqref{eq:bco_objective}, the objective of standard binary-feedback preference optimization (with $\mathcal{D}^+$ and $\mathcal{D}^-$) can be viewed as the minimization of the empirical risk summed over both positive and negative samples:
\begin{align}
    R^+(g) &+ R^-(g) = \mathbb{E}_{(x, y_w) \sim \mathcal{D}^+} [l(g(x, y_w), +1)] \nonumber \\
    &\quad + \mathbb{E}_{(x, y_l) \sim \mathcal{D}^-} [l(g(x, y_l), -1)].
    \label{eq:bco_obj_l}
\end{align}

In the context of personalization, we treat the data from auxiliary users $\mathcal{H}_{\text{aux}}$ as an unlabeled set.
By substituting the negative risk estimator derived in Eq.~\eqref{eq:pu_equality} into the objective, we obtain the raw C-BPO formulation:
\vspace{-0.5mm}
\begin{align}
    \mathcal{L}_{\text{raw}} &= \mathbb{E}_{\mathcal{H}_{\text{tar}}} [l(g, +1)]
     + \frac{1}{\pi_n} \Big( \mathbb{E}_{\mathcal{H}_{\text{aux}}} [l(g, -1)] \nonumber \\
    &\quad - \pi_p \mathbb{E}_{\mathcal{H}_{\text{tar}}} [l(g, -1)] \Big).
    \label{eq:pc_bfo_vanilla}
\end{align}
\vspace{-0.5mm}
This expression enables the model to recover an unbiased preference signal by correcting the negative gradient with information from the positive set during the optimization process.

\paragraph{Assumption Adaptation.}
Applying Eq.~\eqref{eq:pc_bfo_vanilla} requires re-examining the foundational assumptions of PU learning~\cite{hyttinen2013experiment}.
Conventionally, the unlabeled set is required to be a mixture of positive and negative distributions, $p_u(x) = \pi_p p_p(x) + (1-\pi_p) p_n(x)$.
While $\mathcal{H}_{\text{tar}}$ and $\mathcal{H}_{\text{aux}}$ are physically disjoint, we posit that this mixture assumption remains applicable within the shared preference manifold.
Specifically, we treat $\mathcal{H}_{\text{aux}}$ as an implicit mixture where shared features (e.g., generic task knowledge) overlap with the target user's preference space.
This allows us to apply the risk decomposition in Eq.~\eqref{eq:pu_equality} as an estimator for personalized calibration.
A formal discussion on the adaptation of the assumption is provided in Appendix~\ref{appendix:scar_analysis}.



Under this relaxation, the class prior $\pi_p$ is reinterpreted as an empirical correction coefficient $\alpha \in (0, 1)$. 
This coefficient $\alpha$\footnote{We provide an empirical rationale (\S~\ref{appendix:alpha_understand}) and the corresponding estimation procedure  (\S~\ref{appendix:estimation}) for $\alpha$.} quantifies the degree of preference feature overlap between the target user and auxiliary users.

Furthermore, as observed by~\citet{kiryo2017positive}, if the model is highly flexible, the empirical estimator for $R^-(g)$ may become negative during training, leading to overfitting.
To ensure stability and robustness, we adopt the non-negative constraint~\cite{kiryo2017positive}.
The final C-BPO optimization objective is formulated as:
\begin{equation}
    \mathcal{L}_{\text{C-BPO}} = \mathcal{L}_{\text{pos}} + \frac{1}{1-\alpha} \max \left\{ 0, \mathcal{L}_{\text{pure\_neg}} \right\}
    \label{eq:pc_bfo_final}
\end{equation}
where
\begin{align*}
    \mathcal{L}_{\text{pos}} &= \underbrace{\mathbb{E}_{\mathcal{H}_{\text{tar}}} [l(g, +1)]}_{\text{Positive Alignment}}, \\
    \mathcal{L}_{\text{pure\_neg}} &= \underbrace{\mathbb{E}_{\mathcal{H}_{\text{aux}}} [l(g, -1)] - \alpha \mathbb{E}_{\mathcal{H}_{\text{tar}}} [l(g, -1)]}_{\text{Purified Negative Loss}}.
\end{align*}

where the first term ensures the model aligns with the target user's historical preferences, and the second term (Purified Negative Loss) provides a debiased negative signal by explicitly correcting the contamination from common preference features.

\subsection{Adaptation to Imbalanced Preference Data}
\label{sec:imbalance}

In personalized scenarios, the auxiliary user data $\mathcal{H}_{\text{aux}}$ typically far exceeds the target user's data $\mathcal{H}_{\text{tar}}$ in volume.
While leveraging a larger $\mathcal{H}_{\text{aux}}$ may facilitate the modeling of inter-user information, it poses a challenge for the stability of the reference point $\delta$ (\S~\ref{sec:preliminaries}). 



In standard BCO~\cite{jung2025binary}, $\delta$ is estimated by averaging rewards across positive and negative samples jointly within each batch.
While effective in balanced settings, this joint estimator is highly sensitive to sampling ratios; an increased proportion of negative samples can cause $\delta$ to disproportionately drift toward the dominant distribution, thereby undermining its role as a neutral baseline for accurate preference discrimination.    

To mitigate this, we decouple the tracking of reward statistics.
Specifically, we maintain \textbf{independent} Exponential Moving Averages (EMA) for positive and auxiliary rewards.
The calibrated reference point, $\delta_{\text{EMA}}$, is then dynamically computed as the mean of these two decoupled statistics during each optimization step.
This ensures a stable decision boundary that remains invariant to batch-level data imbalance.

\begin{table*}[!t]
\centering
\resizebox{\linewidth}{!}{
\begin{tabular}{lcccccccccc}
\toprule
\multirow{2}{*}{\makecell{\textbf{LaMP Bench.} $\rightarrow$ \\[2pt] \textbf{Method} $\downarrow$}} 
& \multicolumn{2}{c}{\textbf{Abstract Gen.}} 
& \multicolumn{2}{c}{\textbf{Review Writing}} 
& \multicolumn{2}{c}{\textbf{Topic Writing}} 
& \multicolumn{2}{c}{\textbf{News Headline}} 
& \multicolumn{2}{c}{\textbf{Scholarly Title}} \\
\cmidrule(lr){2-3} \cmidrule(lr){4-5} \cmidrule(lr){6-7} \cmidrule(lr){8-9} \cmidrule(lr){10-11}
& R-1 & R-L  & R-1 & R-L  & R-1 & R-L & R-1 & R-L & R-1 & R-L \\
\midrule

\multicolumn{11}{l}{\underline{\textit{Non-Tuned}}} \\
Base Model         & 0.341 & 0.186 & 0.287 & 0.126 & 0.246 & 0.105 & 0.119 & 0.105 & 0.409 & 0.324 \\
RAG                & 0.347 & 0.205 & 0.272 & 0.128 & 0.243 & 0.115 & 0.141 & 0.124 & 0.425 & 0.347 \\
PAG                & 0.344 & 0.186 & 0.256 & 0.125 & 0.262 & 0.107 & 0.118 & 0.102 & 0.372 & 0.289 \\

\midrule
\multicolumn{11}{l}{\underline{\textit{SFT-based}}} \\
TAM                & 0.357 & 0.204 & 0.289 & 0.122 & 0.253 & 0.107 & 0.200 & 0.179 & 0.514 & 0.456 \\
OPPU               & 0.378 & 0.218 & 0.319 & 0.134 & 0.278 & 0.112 & 0.203 & 0.182 & 0.510 & 0.454 \\

\midrule
\multicolumn{11}{l}{\underline{\textit{Preference Opt.}}} \\
CoPE $^\clubsuit$             & \underline{0.392} & \underline{0.239} & \underline{0.335} & \underline{0.146} & \underline{0.281} & \textbf{0.120} & \underline{0.205} & \underline{0.184} & \underline{0.519} & \underline{0.461} \\
KTO $^\dagger$                & 0.370 & 0.229 & 0.298 & 0.126 & 0.269 & 0.109 & 0.191 & 0.173 & 0.491 & 0.431 \\
BCO $^\dagger$                & 0.373 & 0.231 & 0.315 & 0.132 & 0.272 & 0.112 & 0.197 & 0.179 & 0.507 & 0.443 \\

\midrule
\rowcolor{Gray} \textbf{C-BPO (Ours)}$^\dagger$  & \textbf{0.398} & \textbf{0.269} & \textbf{0.353} & \textbf{0.154} & \textbf{0.291} & \underline{0.118} & \textbf{0.215} & \textbf{0.198} & \textbf{0.539} & \textbf{0.481} \\
\bottomrule
\end{tabular}
}
\caption{Results on the LaMP benchmark. R-1 and R-L denote ROUGE-1 and ROUGE-L, respectively. \textbf{Bold} and \underline{underline} mark the best and second-best results. ``$^\dagger$'' denotes the series of binary-feedback preference optimization methods. $^\clubsuit$ indicates the methods implemented based on DPO.}
\vspace{-2mm}
\label{tab:lamp}
\end{table*}

\section{Experiments}

\subsection{Experimental Setup}
\label{sec:exp_setup}

\paragraph{Datasets and Backbone Models.}
We evaluate the effectiveness of C-BPO on personalized generation tasks from the LaMP~\cite{salemi2024lamp} and LongLaMP~\cite{kumar2024longlamp} benchmarks.
Specifically, we select the following representative tasks: News Headline Generation (LaMP-4), Scholarly Title Generation (LaMP-5), Abstract Generation (LongLaMP-2), Review Writing (LongLaMP-3), and Topic Writing (LongLaMP-4).
These datasets provide per-user behavioral history, query inputs, and ground-truth outputs.
Following prior work~\cite{tan2024democratizing, bu-etal-2025-personalized}, we report ROUGE-1 and ROUGE-L scores for all tasks. 
To ensure the robustness of our evaluation, we conduct experiments across several backbone series, including LLaMA~\cite{touvron2023llama, dubey2024llama}, Qwen~\cite{qwen2}, and Mistral~\cite{jiang2023mistral7b}.

\paragraph{Baselines and Implementation Details.}
We compare C-BPO against several categories of baselines: 
(1) \textbf{Retrieval-based methods}: We include \textbf{RAG}~\cite{lewis2020retrieval}, which retrieves user-related histories and appends them to the prompt as in-context evidence, and \textbf{PAG}~\cite{richardson2023integrating}, which incorporates a summarized user profile directly into the prompt to guide the generation process.
(2) \textbf{SFT-based adapters}: We consider \textbf{TAM}~\cite{tan2024democratizing}, a task-specific LoRA module trained on general task data excluding the target user's data, representing a non-personalized upper bound for general task performance. We also compare with \textbf{OPPU}~\cite{tan2024democratizing}, which fine-tunes individual adapters exclusively on each user's historical interaction data via standard supervised learning.
(3) \textbf{Preference-based methods}: We compare against \textbf{CoPE}~\cite{bu-etal-2025-personalized}, which constructs negative samples for each user-specific instance through rejection sampling and optimizes user-specific adapters using DPO~\cite{rafailov2023direct}. Additionally, we evaluate \textbf{KTO}~\cite{ethayarajh2024model} and \textbf{BCO}~\cite{jung2025binary}, which enable preference optimization using binary feedback, allowing for flexible alignment even when the prompts for positive and negative instances differ.

For binary-feedback preference optimization methods (KTO, BCO, and C-BPO), unless otherwise specified, we randomly sample data from other users as the auxiliary set during user-specific training, maintaining a balanced 1:1 ratio with the positive (user-specific) data.
For C-BPO, the calibration coefficient $\alpha$ is pre-estimated prior to training, following the procedure detailed in Appendix~\ref{appendix:estimation}.
To ensure a fair comparison with previous studies, user-specific training is initiated from the TAM checkpoint, and we employ greedy decoding for all evaluations.
Hyperparameters for LoRA modules are kept consistent across all training-based methods.
More details on the experimental setup can be found in Appendix~\ref{appendix:exp_setup}.

\subsection{Main Results}
\label{sec:main_res}
We present the primary experimental results in Table~\ref{tab:lamp} (using \textit{Mistral-7B-Instruct-v0.3}), with extensive evaluations across different base models provided in Appendix~\ref{appendix:res_across_llms} (Figure~\ref{fig:res_across_llms}).
From these results, we derive several key observations:

First, directly applying binary-feedback preference optimization (BPO) methods, such as KTO and BCO, to personalization tasks does not guarantee performance gains.
In several cases, these methods even underperform SFT-based baselines. This phenomenon suggests that \textbf{negative signals from other users can exert a detrimental effect if treated naively}.
As analyzed in \S\ref{sec:motivation}, this is likely due to the preference overlap inherent in personalization scenarios, where generic task-specific knowledge and community-wide preferences are mistakenly penalized.

In contrast, \textbf{C-BPO effectively exploits the negative signals from auxiliary users}, consistently outperforming both SFT-based and standard BPO-based approaches.
This demonstrates that our method successfully leverages inter-user distinctions while avoiding the excessive penalization of shared preferences.
Furthermore, C-BPO achieves competitive or even superior performance compared to the DPO-based method, CoPE.
These results underscore the critical importance of incorporating inter-user information to refine individual preference modeling in LLM personalization.

Furthermore, results presented in Appendix~\ref{appendix:res_across_llms} (Figure~\ref{fig:res_across_llms}) demonstrate that the superior performance of C-BPO generalizes across various backbone LLMs, confirming its robustness and model-agnostic effectiveness.

\section{Analysis}

In this section, we analyze the impact of key components in C-BPO, including the quantity and quality of auxiliary data, the sensitivity of the correction coefficient $\alpha$, and the effectiveness of the EMA-based reference point estimation.

\subsection{The Role of Auxiliary Data}
\label{sec:exp_aux}

\begin{figure}[t]
  \includegraphics[width=\columnwidth]{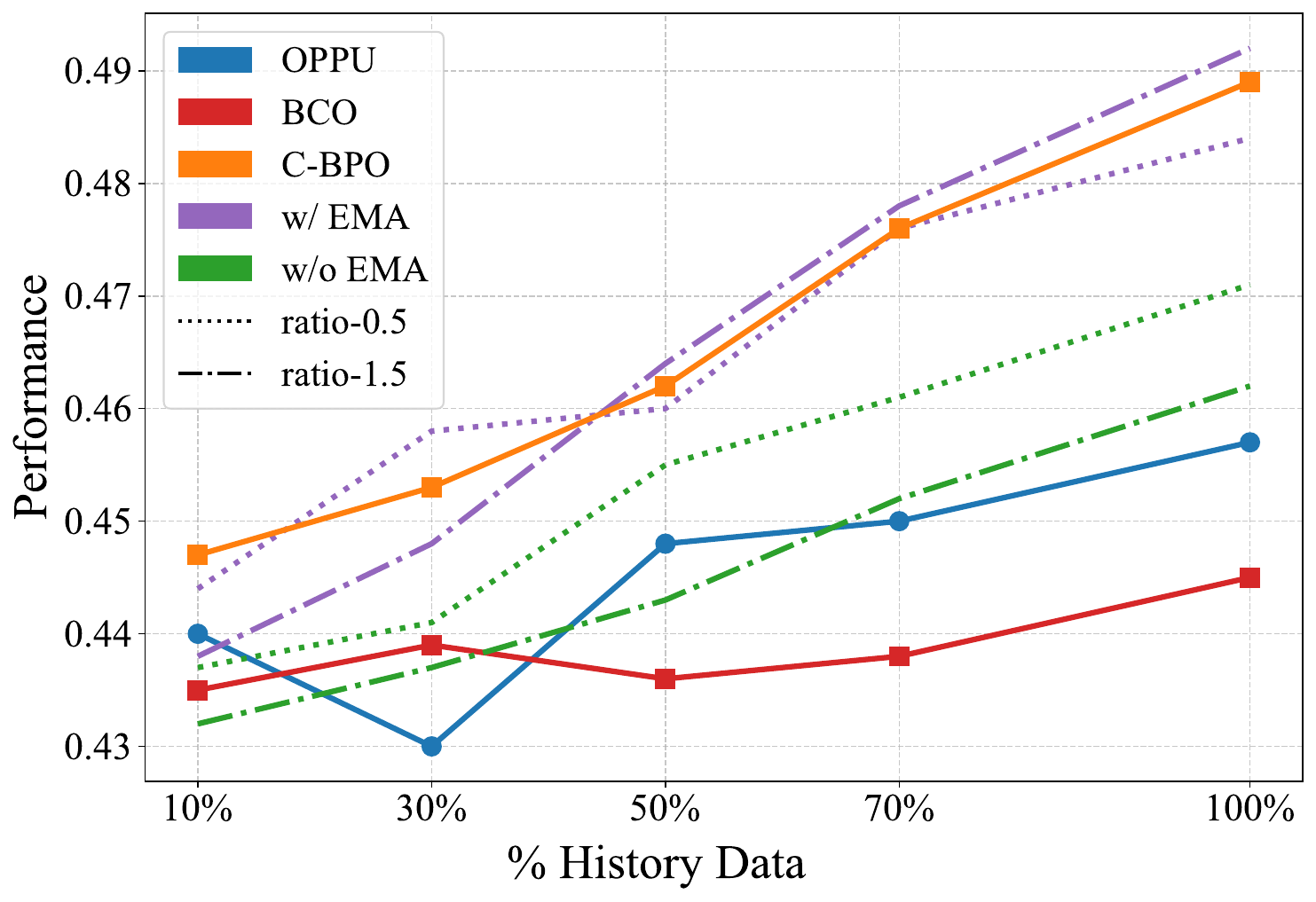}
  \caption{Performance comparison across varying proportions of user historical data ($\mathcal{H}_{\text{tar}}$). $x$ denotes the ratio of auxiliary data to target user data.}
  \label{fig:neg_ratio}
\end{figure}

To further investigate how auxiliary data influences the optimization of personalized LLMs\footnote{In \S~\ref{appendix:log_diff}, we examine how our calibration prevents the erosion of overlap preference during personalization.}, we evaluate the performance of C-BPO under varying proportions of user historical data and different reference point estimation strategies, as illustrated in Figure~\ref{fig:neg_ratio}.
Specifically, we examine the behavior of C-BPO by introducing two variables:
First, we adjust the data ratio $x$ (where $x = \#\mathcal{H}_{\text{aux}} / \#\mathcal{H}_{\text{tar}}$) to $0.5$ and $1.5$, to observe the model's sensitivity to the volume of negative signals.
Second, we compare the standard reference point estimation with our proposed independent EMA-based strategy (\S~\ref{sec:imbalance}).
A more detailed experimental setup is elaborated in \S~\ref{appendix:detailed_aux_data}.

\paragraph{Trade-off in Auxiliary Data Volume.}
We find that increasing the volume of auxiliary data does not always yield monotonic performance gains.
As shown in Figure~\ref{fig:neg_ratio}, when the target user history $\mathcal{H}_{\text{tar}}$ is scarce (e.g., $<50\%$), a higher data ratio ($x=1.5$, \textit{purple dashed line}) yields inferior results compared to the balanced setting ($x=1.0$, \textit{orange solid line}).
Superiority for $x=1.5$ is only achieved after the volume of $\mathcal{H}_{\text{tar}}$ surpasses a certain threshold.
This suggests that effectively distilling inter-user information from $\mathcal{H}_{\text{aux}}$ requires a sufficient amount of $\mathcal{H}_{\text{tar}}$ to provide a strong enough ``positive'' anchor to guide the debiasing process.

\paragraph{Robustness of EMA-based Reference Point Estimation.}
The independent EMA-based estimation proves crucial for handling imbalanced data scenarios.
In the absence of EMA (\textit{green dashed line}), C-BPO exhibits a noticeable performance drop as $x$ deviates from 1.0.
Notably, when $x=1.5$ and $\mathcal{H}_{\text{tar}}$ is limited, the model without EMA even underperforms the SFT-based baseline (OPPU), indicating that a biased reference point can hinder the model's ability to utilize negative signals.
In contrast, the EMA-based strategy (\textit{purple dashed line}) effectively mitigates the sensitivity to $x$ by maintaining a stable decision boundary.
We also observe that with sufficient $\mathcal{H}_{\text{tar}}$, the model becomes more resilient to estimation methods, eventually capturing inter-user signals even without EMA.
This further reinforces that the proposed EMA mechanism is a vital stabilizer for personalized alignment in data-constrained or imbalanced regimes.

\subsection{The Impact of User Uniqueness}
\label{sec:exp_unique}


This section further examines how the degree of user uniqueness (\textbf{preference overlap}) influences personalization performance. Following prior research~\cite{qiu-etal-2025-measuring, qiu2025latent, liu2025llms}, which suggests that the embeddings of users' history effectively reflect user characteristics, we construct different experimental groups by retrieving specific sets of auxiliary data.
Using Euclidean distance in the embedding space relative to $\mathcal{H}_{\text{tar}}$, we curate three distinct configurations: a \textit{Unique} group (retrieving users with the largest distances), a \textit{Non-unique} group (retrieving those with the smallest distances), and a \textit{Random} group.
A more detailed experimental setup is elaborated in \S~\ref{appendix:detailed_user_unique_setup}.
We then evaluate C-BPO and its counterparts across these controlled settings, as illustrated in Figure~\ref{fig:user_uniqueness}\footnote{More results on different datasets are given in \S~\ref{appendix:additional_user_unique_res}}.

\begin{figure*}[t!]
  \centering
   \includegraphics[width=1.0\linewidth]{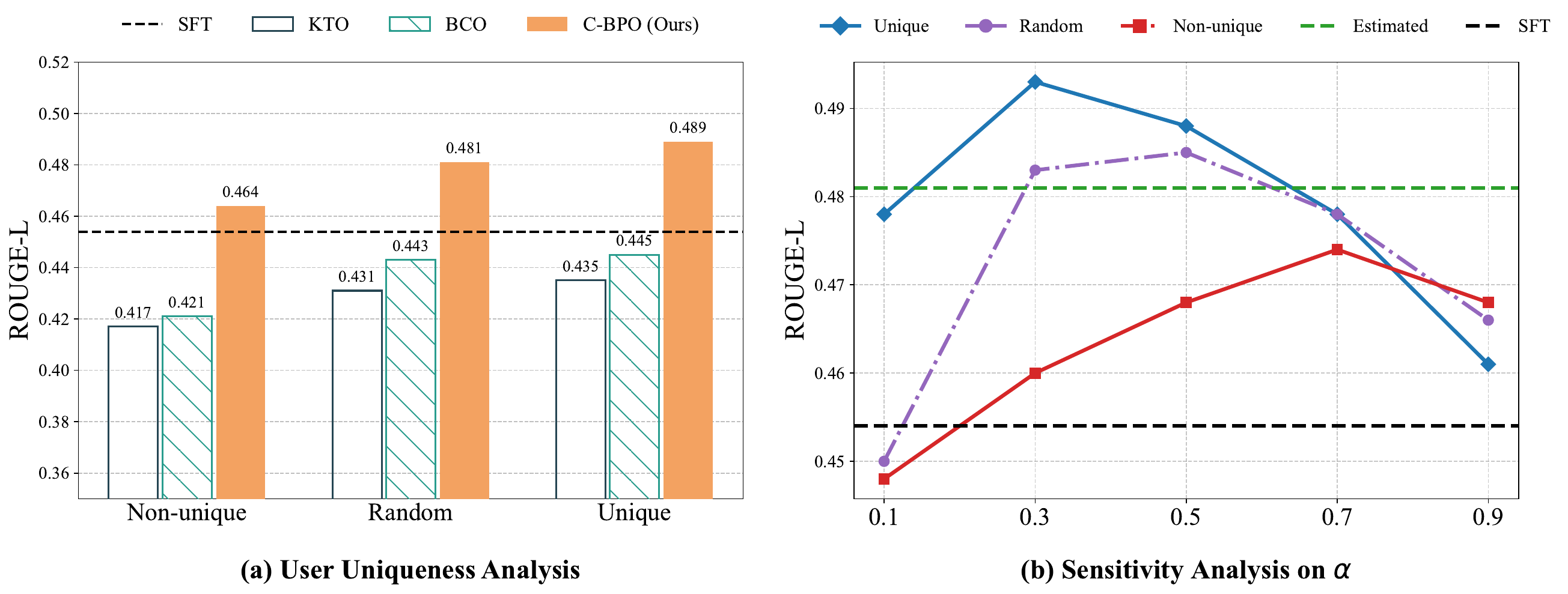}
   \caption{Analysis of user uniqueness and the sensitivity of $\alpha$. (a) Performance across groups retrieved via embedding distances (\textit{Unique}, \textit{Random}, \textit{Non-unique}). (b) C-BPO performance under varying $\alpha$, with ``Estimated'' denoting the performance under the estimation method in \S~\ref{appendix:estimation}.}
   \label{fig:user_uniqueness}
\end{figure*}

\paragraph{Analysis across User Groups.}
As illustrated in Figure~\ref{fig:user_uniqueness} (a), standard BPO-based methods (KTO and BCO) exhibit a significant performance degradation when transitioning from the \textit{Random} to the \textit{Non-unique} setting.
Furthermore, they fail to achieve substantial gains even in the \textit{Unique} setting, consistently underperforming the SFT baseline.
This suggests that traditional BPO approaches are unable to navigate the dense preference overlap in non-unique scenarios and fail to effectively extract useful auxiliary signals when preferences are distinct.
In contrast, C-BPO consistently facilitates preference optimization across all settings by leveraging inter-user information, with the most pronounced improvements observed in the \textit{Unique} group, where individual idiosyncrasies are more prominent and distinct.

\paragraph{Sensitivity and Estimation of the Correction Coefficient $\alpha$.}
Intuitively, the correction coefficient $\alpha$ should scale with the degree of preference overlap.
Figure~\ref{fig:user_uniqueness} (b) confirms this intuition: the \textit{Non-unique} group requires a higher $\alpha$ to filter out common preferences, whereas the optimal performance for the \textit{Unique} group is achieved at a lower $\alpha$ value.
These findings validate the necessity and effectiveness of our preference-calibrated objective.
To address the challenge of hyperparameter tuning, we provid an estimation method based on the embedding of user history in \S~\ref{appendix:estimation}.
As shown by the ``Estimated'' markers in Figure~\ref{fig:user_uniqueness} (b) (for the \textit{Random} group), this heuristic method provides a reliable starting point that closely aligns with the empirically optimal coefficient.

\section{Related Work}

Existing LLM personalization methods primarily fall into three lines: retrieval-based prompting, fine-tuning with user history, and training with auxiliary data. 
Retrieval-based approaches integrate user-specific context into the input via few-shot demonstrations~\cite{brown2020language}, relevant history snippets (RAG, \citealp{mysore2024pearl, salemi2024lamp}), or summarized profiles (PAG, \citealp{richardson2023integrating, guan-etal-2025-survey}). 
While scalable and interpretable, these methods often result in shallow personalization as they rely heavily on the inherent capacity of a powerful LLM to analyze user information within prompt-length constraints. 
Fine-tuning-based methods, in contrast, adapt model parameters through Parameter-Efficient Fine-Tuning (PEFT) for individuals~\cite{tan2024democratizing, tan2024personalized, liu2025llms} or specific user groups~\cite{zhang2025proper}. 
While standard approaches rely solely on historical user data to learn corresponding adapters or embedding prefixes, CoPE~\cite{bu-etal-2025-personalized} further constructs negative samples via rejection sampling to train user-specific adapters using DPO~\cite{rafailov2023direct}, which significantly complicates the overall training and inference pipeline.

Recent advancements have further explored incorporating other users’ information as auxiliary signals, such as using inter-user differences (DPL, ~\citealp[]{qiu-etal-2025-measuring}) or collaborative filtering-inspired retrieval (CFRAG, ~\citealp[]{shi2025retrieval}) to highlight individual uniqueness.
Along this line, DEP~\cite{qiu2025latent} models user difference information within the latent space based on the embedding of the user history data.
Yet, these methods either demand complex reasoning from a large LLM or are confined to latent embedding spaces that may miss fine-grained linguistic nuances.
Moreover, their heavy reliance on sufficient data for high-quality latent representation extraction hinders their ability to generalize or rapidly extend to new users. 

In contrast to the aforementioned methods, our framework migrates binary-feedback preference optimization into the personalization context, relying solely on the target and auxiliary other users' data to construct contrastive signals.
By shifting the focus from high-resource LLM reasoning or holistic embedding alignment to explicit sample-level comparisons, our method effectively captures term- and phrase-level relative information while streamlining the personalization pipeline, enabling more precise alignment with genuine user intent without the necessity for massive datasets or ultra-powerful backbone models.

\section{Conclusion}
This paper introduces C-BPO,  a framework designed for personalizing LLMs through preference-calibrated binary-feedback signals.
Unlike existing methods that rely on isolated user histories, C-BPO captures individuality by treating target user data as positive signals and auxiliary user data as implicit negative feedback.
To address the fundamental challenge of preference overlap, we derive a calibrated objective grounded in Positive-Unlabeled (PU) learning, which purifies negative signals from the auxiliary data.
Comprehensive experiments across five personalized generation tasks demonstrate that C-BPO consistently outperforms competitive baselines, offering a robust, theoretically sound, and scalable solution for aligning LLMs with nuanced individual preferences.

\section*{Acknowledgements}
This work was supported in part by National Natural Science Foundation of China (62476070), Shenzhen Science and Technology Program \seqsplit{(JCYJ20241202123503005, \, GXWD20231128103232001, \,ZDSYS20230626091203008,\, KQTD20240729102154066)}, Department of Science and Technology of Guangdong (2024A1515011540), National Key R\&D Program of China (SQ2024YFE0200592) and Suzhou Science and Technology Program (SYG2025072).

\section*{Limitations}

While C-BPO demonstrates consistent improvements across personalized generation tasks, it possesses several limitations.
First, the framework's effectiveness is sensitive to the calibration of the correction coefficient $\alpha$.
Although we have proposed an initial estimation strategy in Appendix~\ref{appendix:estimation}, more sophisticated and dynamic methods for adaptive parameter estimation remain to be explored.
Second, our method requires access to a centralized auxiliary dataset comprising other users' raw historical data to construct the unlabeled set. It poses privacy risks and deployment hurdles in specific real-world scenarios, such as Federated Learning or on-device personalization, where accessing raw user data is prohibited.
Lastly, our current evaluation is centered on generative benchmarks. Adapting this preference-calibrated objective to broader personalization domains, such as dialogue scenarios or recommendation systems, requires further study to fully validate its cross-domain generalizability.

\section*{Ethical Considerations}

While our work aims to personalize large language models (LLMs), we acknowledge the potential ethical concerns. 
The datasets used in our study may contain biases, which could be reflected in the model’s outputs.
Mitigating such biases is crucial for ensuring fairness. 
Additionally, the use of LLMs may generate offensive or biased content. 
We suggest that practitioners should carefully examine the potential bias before deploying the model in real-world applications.


\bibliography{custom}

\appendix


\crefalias{section}{appendix}
\crefalias{subsection}{appendix}
\crefalias{subsubsection}{appendix}
\startcontents[appendix]
{
    \printcontents[appendix]{ }{0}{\section*
    {Appendix}}
}

\section{The Use of AI Assistants}

Throughout the preparation of this manuscript, large language models were employed exclusively for light stylistic refinement and grammatical adjustment. Furthermore, these tools assisted in generating structural templates for the illustrative diagrams.

\section{The Detailed Difference between BCO and KTO}
\label{appendix:bco_comparison}
\paragraph{KTO} \label{paragraph:kto}
\citet{ethayarajh2024model} proposed alignment framework that trains on binary signal of thumbs-up or thumbs-down collected on a per-sample basis for every unique prompt and completion combination.
Given a dataset of \{ prompt, completion \} pairs with respective binary signals, KTO defines a value function
\begin{align} \label{eq:kto_value}
&v_{KTO}(x, y; \theta) \nonumber \\
&= \begin{cases}
\sigma(r_\theta(x, y) - z_\text{ref}) & \text{if } y \sim y_{\text{desirable}} \mid x \\
\sigma(z_\text{ref} - r_\theta(x, y)) & \text{if } y \sim y_{\text{undesirable}} \mid x,
\end{cases}
\end{align}
where $z_\text{ref}$ is a reference point.
In practice, $z_\text{ref}$ is implemented as
\begin{equation} \label{eq:kto_z_ref}
z_{\text{ref}} = \max \left( 0, \frac{1}{|\mathcal{B}|} \sum_{y' \in \mathcal{B} \setminus y} \log \frac{\pi_\theta(y' \vert x)}{\pi_\text{ref}(y' \vert x)} \right)
\end{equation}
for $(x,y) \in \mathcal{B}$ and $\mathcal{B}=\{(x^{(i)}, y^{(i)})\}_{i=1}^B$ is a batch of samples.

Finally, the loss function of KTO is defined as 
\begin{equation}
\mathcal{L}_{\text{KTO}}(\theta) = \mathbb{E}_{(x,y) \sim \mathcal{D}} \left [ w(y) (1 - v_\text{KTO}(x,y; \theta) \right]
\end{equation}
where the weighting factor $w(y)$ is $\lambda_D$ if $y$ is a completion from thumbs-up dataset and $\lambda_U$ if $y$ is a completion from thumbs-down dataset.

\paragraph{BCO}
\citet{jung2025binary} proposed Binary Classifier Optimization (BCO), a theoretically grounded alignment framework that learns directly from binary feedback.
BCO views alignment from binary signals as training a binary classifier whose logit is the implicit reward induced by the policy–reference log-ratio,
\begin{equation}
r_\theta(x, y) = \beta \log \frac{\pi_\theta(y \mid x)}{\pi_{\text{ref}}(y \mid x)} .
\end{equation}
Under this formulation, minimizing the binary cross-entropy (BCE) loss over thumbs-up and thumbs-down samples provably upper bounds the DPO objective, enabling alignment without explicit preference pairs.

To tighten this bound, BCO introduces a reward-shifting term $\delta$ defined as the average implicit reward of positive and negative samples,
\begin{equation}
\begin{split}
\delta_{\text{BCO}} = \frac{1}{2} \Big( &\mathbb{E}_{(x,y)\sim \mathcal{D}^+}[r_\theta(x,y)] \\
&+ \mathbb{E}_{(x,y)\sim \mathcal{D}^-}[r_\theta(x,y)] \Big).
\end{split}
\end{equation}
where $\mathcal{D}^+$ and $\mathcal{D}^-$ denote thumbs-up and thumbs-down datasets, respectively.
The resulting BCO objective is
\begin{align}
&\mathbb{E}_{(x, y_w) \sim \mathcal{D}^+}
\left[- \log \sigma\!\left(r_\theta(x, y_w) - \delta\right)\right] \nonumber \\
&\quad + \mathbb{E}_{(x, y_l) \sim \mathcal{D}^-}
\left[- \log \sigma\!\left(- (r_\theta(x, y_l) - \delta)\right)\right],
\end{align}
which preserves informative gradients across samples and enables stable, effective alignment purely from binary feedback.

\section{Proofs for Risk Decomposition}
\label{sec:proofs}

In this section, we provide the detailed derivation of the risk decomposition. For completeness, we first restate Lemma~\ref{lem:risk_decomp_main} from the main text before proceeding to its formal proof.

\begin{lemma}[Risk Decomposition, Restated] \label{lem:risk_decomp_appendix}
Let $p_u(x) = \pi_p p_p(x) + \pi_n p_n(x)$ be the marginal distribution of unlabeled data. The negative risk $\pi_n R^-_n(g)$ can be unbiasedly estimated using only positive and unlabeled data distributions as:
\begin{equation}
    \pi_n R^-_n(g) = R^-_u(g) - \pi_p R^-_p(g).
\end{equation}
\end{lemma}

By the definition of the expected risk on unlabeled data $\mathcal{D}_U$, which follows the marginal density $p(x)$, we have:
\begin{align}
    &R^-_u(g) = \int L(g(x), -1) p(x) dx \nonumber \\
    &= \int L(g(x), -1) \left[ \pi_p p_p(x) + \pi_n p_n(x) \right] dx \nonumber \\
    &= \pi_p \int L(g(x), -1) p_p(x) dx \nonumber \\
    &\quad\quad + \pi_n \int L(g(x), -1) p_n(x) dx \nonumber \\
    &= \pi_p R^-_p(g) + \pi_n R^-_n(g). \label{eq:proof_step}
\end{align}
In Eq.~\eqref{eq:proof_step}, the first term $\pi_p R^-_p(g)$ represents the risk contribution from positive samples that are mistakenly treated as negative in the unlabeled set. Rearranging the terms, we obtain the estimator for the true negative risk: $\pi_n R^-_n(g) = R^-_u(g) - \pi_p R^-_p(g)$.

\section{Review of PU Learning}
\label{sec:pu_learning}

In this section, we provide a more formal background on Positive-Unlabeled (PU) learning~\cite{kiryo2017positive, bekker2020learning, wang2023pue} and the derivation of the unbiased risk estimator used in our framework.

\paragraph{From PN to PU Risk}
In an ideal scenario where both positive and negative preference data are available, a model $g$ parametrized by $\theta$ is optimized by minimizing the expected risk:
\begin{equation}
    R(g) = \pi R^+_p(g) + (1-\pi) R^-_n(g),
    \label{eq:app_pn_risk}
\end{equation}
where $\pi = P(y=1)$ is the class prior, while $R^+_p(g) = \E_{x \sim p_p}[L(g(x), +1)]$ and $R^-_n(g) = \E_{x \sim p_n}[L(g(x), -1)]$ denote the expected positive and negative risks.
In the PU learning setting, a representative negative set $\mathcal{D}_N$ is unavailable. Instead, we possess a positive set $\mathcal{D}_P$ and an unlabeled set $\mathcal{D}_U$ drawn from the marginal density $p(x) = \pi p_p(x) + (1-\pi) p_n(x)$.

\paragraph{Unbiased Risk Estimator}
The key challenge in PU learning is to estimate the negative risk $R^-_n(g)$ without negative samples.
Utilizing the marginal density relationship, we can express the expected negative loss on the unlabeled distribution as a mixture:
\begin{equation}
    R^-_u(g) = \pi R^-_p(g) + (1-\pi) R^-_n(g),
\end{equation}
where $R^-_u(g) = \E_{x \sim p(x)}[L(g(x), -1)]$ and $R^-_p(g) = \E_{x \sim p_p}[L(g(x), -1)]$. By rearranging this identity, we obtain an unbiased estimator for the weighted negative risk:
\begin{equation}
    (1-\pi) R^-_n(g) = R^-_u(g) - \pi R^-_p(g).
\end{equation}
Substituting this back into Eq.~\eqref{eq:app_pn_risk}, the total risk for PU learning becomes:
\begin{equation}
    R_{PU}(g) = \pi R^+_p(g) + R^-_u(g) - \pi R^-_p(g).
\end{equation}


\section{Empirical Adaptation of SCAR in Preference Spaces}
\label{appendix:scar_analysis}

\subsection{Assumption Adaptation}
\label{appendix:assumption_ada}

A potential concern in applying Positive-Unlabeled (PU) learning to personalization is the validity of the \textbf{Selected Completely At Random (SCAR)} assumption.
In its classic form, SCAR posits that labeled positive instances are sampled randomly from the underlying positive distribution, implying that the unlabeled set $\mathcal{D}_{U}$ serves as a representative mixture of both positive and negative classes.
While this assumption is traditionally applied to classification tasks with well-defined class separability, it requires a nuanced reinterpretation within the context of high-dimensional, personalized preference alignment.

In our personalization context, although $\mathcal{H}_{\text{aux}}$ (historical data from auxiliary users) does not physically encompass the target user's specific historical data, we argue that the preference distributions across different users are fundamentally entangled in the latent feature space.

\paragraph{Preference Feature-Space Decomposition.}
For the sake of simplicity, we assume that the representation of a preference sample $x$ in the latent manifold can be decomposed into two orthogonal components: a \textit{general} component $x_{gen}$ (representing \textit{generic task-specific knowledge} and \textit{community-wide preferences}) and a \textit{specific} component $x_{spec}$ (representing \textit{individual-specific preference}).
Formally, we define the feature mappings as:
\begin{align}
    \text{Feature}(U) &\approx w_{u,1} \cdot \Phi_{\text{gen}} + w_{u,2} \cdot \Phi_{\text{spec\_neg}}, \\
    \text{Feature}(P) &\approx w_{p,1} \cdot \Phi_{\text{gen}} + w_{p,2} \cdot \Phi_{\text{spec\_pos}}.
\end{align}

\paragraph{The Mechanism of Bias and Correction.}
The bias arises because $\mathcal{H}_{\text{aux}}$ and $\mathcal{H}_{\text{tar}}$ overlap significantly in the $\Phi_{\text{gen}}$ subspace. If we were to naively penalize all samples in $\mathcal{H}_{\text{aux}}$ (i.e., minimizing $\mathbb{E}_{\mathcal{H}_{\text{aux}}}[l^-]$), the model would inevitably receive negative gradients for the general features $\Phi_{\text{gen}}$.
This leads to the undesirable effect of the model becoming less helpful or coherent simply because it overfits to the target user's specific taste at the expense of general capabilities.

By reinterpreting the class prior $\pi$ as a correction coefficient $\alpha$, our objective $\mathbb{E}_{\mathcal{H}_{\text{aux}}}[l^-] - \alpha \mathbb{E}_{\mathcal{H}_{\text{tar}}}[l^-]$ functions as a gradient counter-balancing mechanism:
\begin{itemize}
    \item $\mathbb{E}_{\mathcal{H}_{\text{aux}}}[l^-]$ generates a negative gradient on $\Phi_{\text{gen}} + \Phi_{\text{spec\_neg}}$.
    \item $-\alpha \mathbb{E}_{\mathcal{H}_{\text{tar}}}[l^-]$ generates a \textit{positive} gradient on $\Phi_{\text{gen}} + \Phi_{\text{spec\_pos}}$.
\end{itemize}
Since $\mathcal{H}_{\text{tar}}$ and $\mathcal{H}_{\text{aux}}$ are ``entangled'' within the $\Phi_{\text{gen}}$ manifold, the positive gradient from the term $-\alpha \mathbb{E}_{\mathcal{H}_{\text{tar}}}[l^-]$ precisely offsets the erroneous negative gradient produced by $\mathcal{H}_{\text{aux}}$ on the shared features.

\paragraph{Conclusion on SCAR Adaptation.}
Under this interpretation, the SCAR assumption holds not on the physical identity of the samples, but on the \textbf{shared feature manifold}. The target user's positive samples $P\in\mathcal{H}_{\text{tar}}$ can be viewed as being ``randomly selected'' from the broader distribution of ``high-quality/preferred features'' available in the unlabeled pool $\mathcal{H}_{\text{aux}}$.
\textbf{The coefficient $\alpha$ thus reflects the density of these shared features $\Phi_{\text{gen}}$ relative to the total auxiliary set}, providing a theoretically grounded way to calibrate the debiasing strength without compromising the model's fundamental performance.
We also provide a heuristic way to estimate the correction coefficient $\alpha$ in Appendix~\ref{appendix:estimation}.

\subsection{Intuitive Understanding of the Correction Coefficient $\alpha$}
\label{appendix:alpha_understand}

To provide a straightforward interpretation of $\alpha$, we establish a conceptual mapping between the class prior $\pi_p$ in PU learning and our correction coefficient.
In binary classification, $\pi_p$ represents the mixing proportion of positive instances within the unlabeled set.
In our personalized generation context, this generalizes to the density of preference overlap: $\alpha$ quantifies the extent to which the auxiliary user data $\mathcal{H}_{\text{aux}}$ contains features that are intrinsically aligned with the target user's preference manifold.
While a high $\pi_p$ suggests a label-contaminated unlabeled set, a high $\alpha$ indicates that the target user shares significant commonalities (e.g., generic task-specific knowledge and community-wide preferences) with the broader population, necessitating stronger calibration to avoid suppressing these shared positive traits.

\begin{table*}[t]
\centering
\small
\renewcommand{\arraystretch}{1.1}
\resizebox{\textwidth}{!}{%
\begin{tabular}{l|cc|cc}
\toprule
\multirow{2}{*}{\textbf{Task}} & \multicolumn{2}{c|}{\textbf{Other Users}} & \multicolumn{2}{c}{\textbf{Target Users}} \\
 & \#Profile & Output Length & \#Profile & Output Length \\
\midrule
Abstract Generation        & 31,808 & 233.1 $\pm$ 117.5 & 1,296.7 $\pm$ 446.4 & 210.5 $\pm$ 92.8 \\
Review Writing             & 19,649 & 407.2 $\pm$ 299.5 & 759.3 $\pm$ 324.2   & 511.8 $\pm$ 294.2 \\
Topic Writing              & 21,119 & 358.3 $\pm$ 316.9 & 260.6 $\pm$ 314.0   & 358.3 $\pm$ 255.4 \\
News Headline Generation   & 7,275  & 15.5 $\pm$ 6.0    & 270.1 $\pm$ 182.1   & 18.6 $\pm$ 5.2 \\
Scholarly Title Generation & 16,076 & 17.9 $\pm$ 6.1    & 444.0 $\pm$ 121.6   & 16.4 $\pm$ 5.8 \\
\bottomrule
\end{tabular}
}
\caption{The statistics of the used dataset.}
\label{tab:dataset_sta}
\end{table*}

\subsection{Estimation of the Correction Coefficient $\alpha$}
\label{appendix:estimation}

The correction coefficient $\alpha$ in our framework shares a fundamental conceptual link with the class prior $\pi_p$ in Positive-Unlabeled (PU) learning (\S~\ref{sec:pu}).
Specifically, $\alpha$ quantifies the degree of preference overlap, representing the proportion of auxiliary data that functionally serves as positive signal relative to the target user.
Inspired by the seminal Elkan-Noto method~\cite{elkan2008learning, jain2020class}, which estimates the class prior by evaluating the labeling propensity of unlabeled instances, we design a prior estimation scheme for $\alpha$ based on latent representations.

Given that historical embeddings effectively capture distinct user characteristics~\cite{liu2025llms, qiu2025latent}, we formalize the estimation as follows:

\begin{enumerate}

\item \textbf{Training the Proxy Classifier:} We train a probabilistic classifier $g(x)$ to distinguish between the target user history $\mathcal{H}_{\text{tar}}$ (proxy class 1) and the auxiliary user history set $\mathcal{H}_{\text{aux}}$ (proxy class 0) in the embedding space. This classifier learns to identify features indicative of the target user's specific preference manifold.

\item \textbf{Estimation of Propensity Score ($c$):} Following the SCAR assumption that labeled target instances are representative of the underlying positive distribution, we use a held-out validation set $h_t \subset \mathcal{H}_{\text{tar}}$ to estimate the constant labeling propensity $c$:
\begin{equation}
    \hat{c} = \frac{1}{|h_t|} \sum_{x \in h_t} g(x).
\end{equation}

\item \textbf{Calculation of $\alpha$:} With the estimate $\hat{c}$ and the trained classifier $g(x)$, the correction coefficient $\alpha$ is derived by averaging the predictions over the auxiliary set $\mathcal{H}_{\text{aux}}$, representing the density of "target-like" preferences within the broader population:
\begin{equation}
    \hat{\alpha} = \frac{1}{|\mathcal{H}_{\text{aux}}| \cdot \hat{c}} \sum_{x \in \mathcal{H}_{\text{aux}}} g(x).
\end{equation}
\end{enumerate}

This estimation provides a theoretically grounded heuristic for $\alpha$ before training, enabling the framework to adapt to varying degrees of user uniqueness as empirically validated in \S~\ref{sec:exp_unique}.

\begin{figure*}[t!]
  \centering
   \includegraphics[width=1.0\linewidth]{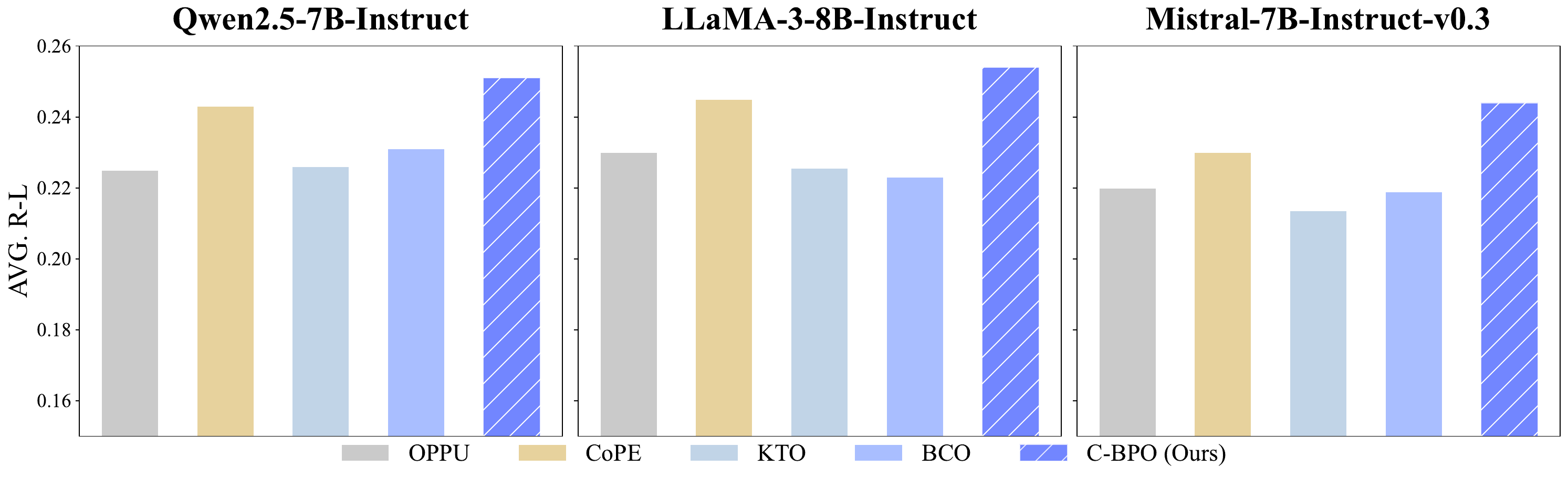}
   \caption{Average performance across 5 tasks for various LLMs.}
   \label{fig:res_across_llms}
\end{figure*}

\section{Details of Experimental Setup}
\label{appendix:exp_setup}

\subsection{Dataset}

We evaluate our method on several text generation tasks from the \textbf{LaMP}~\cite{salemi2024lamp} and \textbf{LongLaMP}~\cite{kumar2024longlamp} benchmarks, which are designed to test the personalization capabilities of LLMs across diverse contexts.

Following the established protocols from previous work~\cite{tan2024democratizing, bu-etal-2025-personalized}, we select the following representative tasks:

\begin{description}
    \item[LaMP-4: News Headline Generation.] This task requires the model to generate concise headlines for news articles. It emphasizes capturing an author's distinctive journalistic style based on their historical article-title pairs.
    \item[LaMP-5: Scholarly Title Generation.] Models must generate titles for scholarly abstracts. Success depends on reflecting the author's specific academic writing style as evidenced in their publication history.
    \item[LongLaMP-2: Abstract Generation.] This task evaluates the model's proficiency in generating scientific abstracts from titles and keywords. It requires emulating characteristic academic phrasing and domain-specific terminology from a user's previous publications.
    \item[LongLaMP-3: Review Writing.] The objective is to generate product reviews based on specifications. The model must reflect the user's evaluative style and subjective perspective captured in their review history.
    \item[LongLaMP-4: Topic Writing.] This task involves generating Reddit post content from provided summaries. The model must maintain the unique, often informal, writing style characteristic of individual users across their historical posts.
\end{description}

The detailed statistics of the datasets are provided in Table~\ref{tab:dataset_sta}.

\subsection{Implementation Details}
\label{appendix:detailed_implement}

We provide the additional implementation details for C-BPO and all baseline methods below:

\begin{itemize}
    \item \textbf{General Training Configuration:}  
    In alignment with prior work~\cite{bu-etal-2025-personalized}, all methods involving a Supervised Fine-Tuning (SFT) stage are optimized using AdamW~\cite{loshchilovdecoupled} with a weight decay of 0.01. We employ a linear learning rate scheduler with a warm-up ratio of 0.1. SFT is conducted for 2 epochs, with learning rates set to $1 \times 10^{-4}$ for LongLaMP and $1 \times 10^{-5}$ for LaMP. For LoRA-based adaptation, we configure the rank $r=8$ and scaling factor $\alpha=16$. All experiments are executed on NVIDIA L40S GPUs.
    
    \item \textbf{Configuration for Binary-Feedback Preference Optimization:}  
    For methods based on binary-feedback preference optimization (i.e., KTO, BCO, and C-BPO), we utilize data from auxiliary users as the negative set during user-specific training.
    Unless otherwise specified, we maintain a 1:1 ratio between the positive (target user) and negative (auxiliary users) samples, where the auxiliary data are randomly drawn from the histories of other users.
    These methods are trained for 3 epochs with a unified learning rate of $1 \times 10^{-6}$.
    The hyperparameter $\alpha$ in C-BPO is pre-estimated prior to training, following the procedure detailed in Appendix~\ref{appendix:estimation}.
\end{itemize}

\subsection{Experimental Settings for Auxiliary Data Analysis}
\label{appendix:detailed_aux_data}

To further investigate how auxiliary data and reference point estimation strategies influence the optimization of personalized LLMs, we conduct a series of controlled experiments.
We specifically select the top 40 users with the most extensive historical data from the LaMP benchmark's test user set to ensure sufficient data for scaling analysis.
During training, we systematically vary the proportion of utilized historical data (controlled by percentage) to observe model performance across diverse data-density regimes.

Two primary dimensions are explored in this analysis:
\begin{itemize}
    \item \textbf{Sensitivity to Auxiliary Data Volume:} We adjust the data ratio $x$ (defined as $x = |\mathcal{H}_{aux}|/|\mathcal{H}_{tar}|$) to values of 0.5 and 1.5. By dynamically controlling the proportion of training data, we observe the model's sensitivity to the volume of negative signals relative to target user data.
    
    \item \textbf{Impact of Reference Point Calibration:} Across varying data ratios $x$, we evaluate the effectiveness of our proposed independent EMA-based strategy (\S~\ref{sec:imbalance}). This comparison highlights the framework's robustness under different degrees of data imbalance.
\end{itemize}

The corresponding experimental results are illustrated in Figure~\ref{fig:neg_ratio}.

\begin{figure*}[t!]
  \centering
   \includegraphics[width=1.0\linewidth]{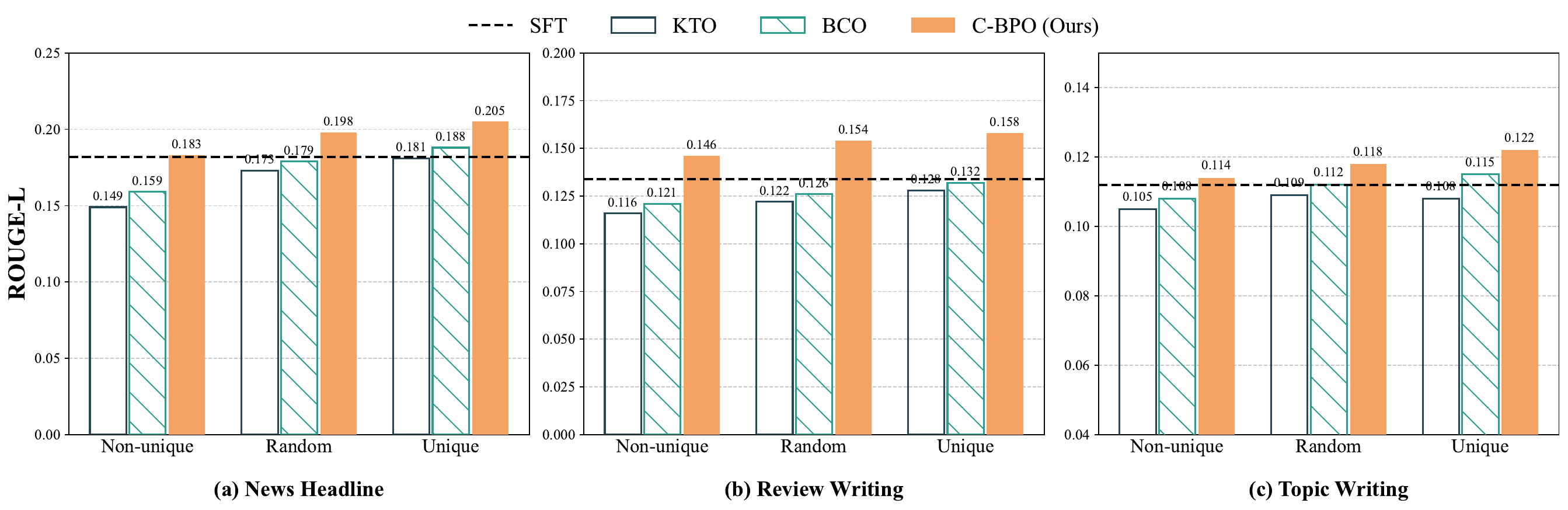}
   \caption{User uniqueness analysis across different tasks.}
   \label{fig:user_uniqueness_across_taks}
\end{figure*}

\begin{figure}[t]
  \includegraphics[width=\columnwidth]{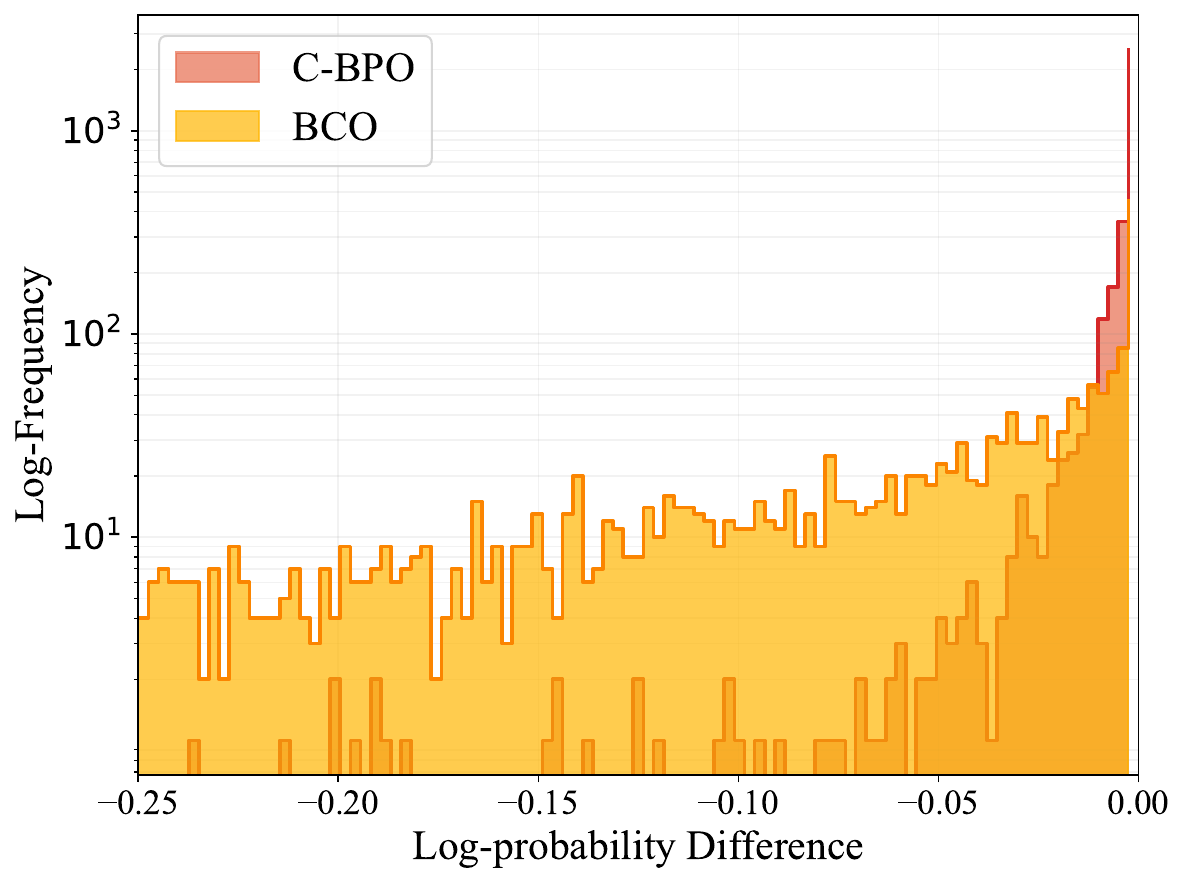}
  \caption{Token-level log-probability shift on auxiliary data. C-BPO effectively mitigates the over-penalization of shared preferences compared to the BCO.}
  \label{fig:logp_diff}
\end{figure}

\subsection{Experimental Settings for User Uniqueness Analysis}
\label{appendix:detailed_user_unique_setup}

To investigate the impact of \textbf{preference overlap} (\S~\ref{sec:motivation}) on personalized training, we follow prior research \cite{qiu2025latent, liu2025llms} which suggests that embedding-based representations of user history effectively reflect individual characteristics.
We construct distinct experimental groups by retrieving auxiliary data based on their semantic proximity to the target user.
Specifically, we utilize \textbf{BGE-M3}~\cite{chen2024bge} as the embedding model to map each historical data point into a high-dimensional vector space.
Using the \textbf{Euclidean distance} between the embeddings of the target user's history $\mathcal{H}_{tar}$ and the historical data of other users, we retrieve an equal volume of samples to form $\mathcal{H}_{aux}$.
Based on this distance metric, we define three comparative groups:
\begin{itemize}
    \item \textbf{\textit{Non-unique group}:} Comprised of auxiliary data with the smallest Euclidean distance to $\mathcal{H}_{tar}$, representing a high degree of preference overlap.
    \item \textbf{\textit{Unique group}:} Comprised of auxiliary data with the largest Euclidean distance, representing significant user divergence.
    \item \textbf{\textit{Random group}:} Formed by randomly sampling auxiliary data to serve as a baseline reference.
\end{itemize}

The comparative results across these groups are illustrated in Figure~\ref{fig:user_uniqueness} and Figure~\ref{fig:user_uniqueness_across_taks}.

\section{Additional Experimental Results}
\label{appendix:additional_res}

\subsection{Additional Results across Different Backbone LLMs}
\label{appendix:res_across_llms}
To evaluate the generalizability of our approach, we perform experiments on multiple backbone LLMs. The results in Figure~\ref{fig:res_across_llms} demonstrate that the superiority of C-BPO is consistent across different architectures, validating its robustness and model-agnostic nature.

\subsection{Additional Results for User Uniqueness}
\label{appendix:additional_user_unique_res}

We further extend our analysis by evaluating user grouping across different datasets, following the experimental setup detailed in \S~\ref{sec:exp_unique}, with results presented in Figure~\ref{fig:user_uniqueness_across_taks}. The empirical findings consistently align with the primary observations discussed in \S~\ref{sec:exp_unique}.

\subsection{Analysis on Preservation of Overlap Preference} 
\label{appendix:log_diff}

To empirically verify that C-BPO effectively mitigates the excessive penalization of auxiliary data, we analyze the shift in token-level log-probabilities before and after personalization.

\paragraph{Experimental Setup.}
We focus on a ``Non-Unique User'' group identified via the embedding-based clustering strategy described in \S~\ref{appendix:detailed_user_unique_setup}.
Within this group, users exhibit high behavioral similarity. We first fine-tune a base LLM on the entire group to obtain a general preference model, $\pi_{\text{gen}}$.
Subsequently, we randomly select a target user $\mathcal{H}_{\text{tar}}$ and treat the remaining users as the auxiliary set $\mathcal{H}_{\text{aux}}$.
We then perform personalized training using both the standard BPO (the baseline) and our proposed C-BPO, resulting in two personalized models: $\pi_{\text{BPO}}$ and $\pi_{\text{C-BPO}}$.

\paragraph{Evaluation Metric.}
We assess the distribution shift by measuring the token-level log-probability difference on the auxiliary data $\mathcal{H}_{\text{aux}}$:
\begin{equation}
\Delta \log P = \log \pi_{\text{pers}}(y|x) - \log \pi_{\text{gen}}(y|x),
\end{equation}
where $\pi_{\text{pers}} \in \{\pi_{\text{BPO}}, \pi_{\text{C-BPO}}\}$.
This metric quantifies the extent to which personalization alters the model's confidence in ``overlap preference'' knowledge shared between the target and auxiliary users.

\textbf{Observations.}
As illustrated in Figure~\ref{fig:logp_diff}, the standard BCO objective leads to a significant decline in the log-probabilities of auxiliary tokens.
Given the high similarity within the user group, this decline indicates that the standard objective erroneously treats shared linguistic patterns and common preferences as negative signals, suppressing them during optimization.
In contrast, C-BPO maintains a higher and more stable probability distribution on $\mathcal{H}_{\text{aux}}$.
This result confirms that our preference-calibrated objective effectively ``protects'' shared knowledge, ensuring that the model differentiates the user only on truly idiosyncratic traits without compromising the foundational commonalities of the user community.

\end{document}